\documentclass[fancyhdr]{article}

\usepackage{authblk}
\usepackage[utf8]{inputenc}
\usepackage[T1]{fontenc}
\usepackage{pslatex}
\usepackage[english]{babel}
\usepackage{fancyhdr}
\usepackage{hyperref}
\usepackage{amsmath,amssymb,amsfonts,amsthm}
\usepackage{graphicx}
\usepackage{amsmath}
\usepackage{amsthm}
\usepackage{amsfonts}
\usepackage{mathtools}
\usepackage{amssymb}
\usepackage{algorithm}
\usepackage{algorithmic}
\usepackage{microtype}
\usepackage{rotating}
\usepackage{todonotes}
\usepackage{tikz}
\usepackage{tikz-qtree}
\usepackage{xparse}
\usepackage{appendix}
\usepackage{url}
\usepackage{tabularx}
\usepackage{paralist}
\usepackage{booktabs}
\usepackage{multirow}
\usepackage{subcaption}
\usepackage{cancel}
\usetikzlibrary{calc,shapes.callouts,shapes.arrows}
\usetikzlibrary{automata}
\usetikzlibrary{positioning}
\usetikzlibrary{decorations.pathreplacing}
\usetikzlibrary{decorations.pathmorphing}
\usetikzlibrary{arrows}
\usetikzlibrary{shapes}

\newcommand{\Z}{\mathbb{Z}}
\newcommand{\F}{\mathbb{F}}

\newtheorem*{problem*}{Problem}

\providecommand{\keywords}[1]{\textbf{\textit{Keywords }} #1}

\begin{document}

\title{Monotone but Exciting: On Evolving Monotone Boolean Functions with High Nonlinearity}

\author[1,2]{Claude Carlet}
\author[3]{Marko \v{C}upi\'c}
\author[3]{Marko \DH urasevic}
\author[3]{Domagoj Jakobovic}
\author[4]{Luca Mariot}
\author[3,5]{Stjepan Picek}

\affil[1 ]{{\normalsize University of Paris 8, 2 rue de la libert\'{e}, 93526 Saint-Denis Cedex, France}}

\affil[2 ]{{\normalsize University of Bergen, Bergen, Norway}

    {\small \texttt{claude.carlet@gmail.com}}}

\affil[3 ]{{\normalsize Faculty of Electrical Engineering and Computing, University of Zagreb, Unska 3, Zagreb, Croatia} \\

{\small \texttt{marko.cupic@fer.hr, marko.durasevic@fer.hr, domagoj.jakobovic@fer.hr}}}

\affil[4 ]{{\normalsize Semantics, Cybersecurity and Services Group, University of Twente, Drienerlolaan 5, 7522 NB Enschede, The Netherlands} \\
	
	{\small \texttt{l.mariot@utwente.nl}}}

\affil[5 ]{{\normalsize Digital Security Group, Radboud University, Postbus 9010, 6500 GL Nijmegen, The Netherlands} \\
	
	{\small \texttt{stjepan@computer.org}}}
	
\maketitle

\begin{abstract}
Monotone Boolean functions are a structurally important class of Boolean functions, but their restricted form imposes strong limitations on achievable nonlinearity. In this paper, we investigate whether evolutionary computation can evolve monotone Boolean functions with high nonlinearity, both in the balanced and imbalanced settings. We consider three solution encodings: the standard truth table representation, a balanced truth table encoding that preserves Hamming weight, and a symbolic tree-based genetic programming representation. To guide the search toward monotone increasing functions, we introduce a non-monotonicity penalty and combine it with fitness functions targeting balancedness and nonlinearity. Experimental results are reported for dimensions from $n=5$ to $n=14$. The results show that evolutionary search can discover monotone Boolean functions with nonlinearities clearly exceeding those of majority functions, and in several cases approaching the best currently known values for monotone functions. At the same time, the experiments reveal substantial differences between encodings: the balanced truth table encoding performs poorly for larger dimensions, while the standard truth table and genetic programming encodings remain competitive, with genetic programming becoming especially relevant in the largest tested dimensions.
\end{abstract}

\keywords{Boolean functions, Monotone Functions, Nonlinearity}

\section{Introduction}
\label{sec:intro}

Boolean functions are mathematical objects used in diverse applications, and as such, we require Boolean functions with diverse properties and structures.
One especially interesting subclass is that of monotone Boolean functions. These functions satisfy a simple and natural order constraint: switching an input bit from 0 to 1 can never force the output to decrease. 
Monotone Boolean functions are useful whenever adding more ``positive evidence'' should never make the output flip the wrong way.
Monotone functions arise naturally in threshold phenomena~\cite{kalai2006threshold}, decision processes~\cite{odonnell2014analysis}, system biology~\cite{pauleve2022booleannetworks}, and combinatorial models~\cite{korshunov2003monotone}, and include important families such as monomials, threshold functions, and majority functions. At the same time, this structural restriction significantly limits their spectral behavior: known theoretical bounds show that monotone Boolean functions must remain much closer to affine functions than unrestricted Boolean functions, and even standard representatives such as majority functions fall well below the best-known nonlinearities in the unrestricted setting~\cite{carlet_2021}.

This tension between structural simplicity and spectral complexity makes monotone Boolean functions a compelling object of study. While the nonlinearity of several explicit monotone families is known, and upper bounds for the class have been established, much less is known about how close one can get to these limits in practice, especially when searching beyond classical constructions. At the same time, evolutionary and metaheuristic methods have proved highly successful in the design of Boolean functions with desirable properties, including balanced, bent, and highly nonlinear functions~\cite{Djurasevic2023}. However, to the best of our knowledge, these methods have not yet been systematically explored for the evolution of monotone Boolean functions.

In this paper, we investigate whether evolutionary computation can effectively discover monotone Boolean functions with high nonlinearity. We consider both balanced and imbalanced settings, examine several solution encodings, and design fitness functions that combine monotonicity enforcement with spectral optimization. By doing so, we aim to better understand the search landscape of monotone Boolean functions and to assess how close evolutionary approaches can come to the best-known bounds and benchmark constructions such as threshold and majority functions.

Our main contributions are:
\begin{compactenum}
    \item This is the first systematic study considering the design of monotone Boolean functions with high nonlinearity with evolutionary algorithms. 
    \item We consider the evolutionary design of both balanced and imbalanced monotone functions, comparing three encodings and several fitness variants. The results show that the nonlinearity of evolved monotone functions outperforms that of majority functions.
\item We identify and discuss which encodings remain effective as the number of input variables increases.
\end{compactenum}

\section{Background}
\label{sec:preliminaries}

We denote by $\mathbb F_2$ the finite field with two elements $0,1$, equipped with XOR (sum) and logical AND (multiplication), respectively. The $n$-dimensional vector space over $\mathbb F_2$ is denoted by $\mathbb F_2^n$, consisting of all $2^n$ binary vectors of length $n$. Given $a, b \in \mathbb F_2^n$, their inner product equals $a\cdot b = \bigoplus_{i=1}^{n} a_{i}b_{i}$ in $\mathbb F_{2}^n$. A Boolean function of $n$ variables is a mapping $f: \F_2^n \to \F_2$. 
Finally, we equip $\mathbb{F}_2^n$
with the usual coordinate-wise partial order:
\begin{equation}
    u \preceq v
\quad \Longleftrightarrow \quad
u_i \le v_i \text{ for every } i \in \{1,\dots,n\}.
\end{equation}

\subsection{Representations and Properties of Boolean Functions}

\subsubsection{Truth Table Representation.}
A Boolean function $f: \F_2^n \to \F_2$ can be uniquely represented by using its truth table. The truth table (TT) of a Boolean function $f$ is the list of pairs $(x, f(x))$ of input vectors $x \in \F_2^n$ and function outputs $f(x) \in \F_2$. Once a total order has been fixed on the input vectors of $\F_2^n$ (most commonly, the lexicographic order), the truth table can be represented by the simple output vector of the function, of length $2^n$.
    
\subsubsection{Walsh-Hadamard Transform.}
The Walsh-Hadamard transform $W_{f}: \F_2^n \to \Z$ is another unique representation of a Boolean function $f$. The Walsh-Hadamard transform measures the correlation between $f$ and the linear functions $a\cdot x$, for all $a \in \mathbb F_2^n$:
\begin{equation}
W_{f} (a) = \sum\limits_{x \in \mathbb{F}_{2}^{n}} (-1)^{f(x) \oplus a\cdot x},
\end{equation}
with the sum calculated in ${\mathbb Z}$. The Walsh-Hadamard transform is an involution up to a normalization by a constant. As such, one can retrieve the truth table representation of $f$ from the spectrum of its Walsh-Hadamard coefficients $W_f(a)$. The Walsh-Hadamard transform allows easy characterization of several cryptographic properties, particularly balancedness and nonlinearity.

\subsubsection{Balancedness}
A Boolean function is balanced if it has the same number of zeros and ones in its truth table. Alternatively, it is balanced if $W_{f} (0) = 0$~\cite{carlet_2021}.

\subsubsection{Nonlinearity}
The minimum Hamming distance between a Boolean function $f$ and all affine functions is the nonlinearity of $f$, which is calculated from the Walsh-Hadamard spectrum as follows~\cite{carlet_2021}:
\begin{equation}
\label{eq:nonlinearity}
nl_{f} = 2^{n - 1} - \frac{1}{2}\max_{a \in \mathbb{F}_{2}^{n}} \left \{ |W_{f}(a)| \right \}.
\end{equation}
Every $n$-variable Boolean function $f$ satisfies the covering radius bound:
\begin{equation}
\label{eq_boolean_covering}
    nl_{f} \leq 2^{n-1}-2^{\frac n 2 - 1}.
\end{equation}
Eq.~\eqref{eq_boolean_covering} cannot be tight when $n$ is odd (since $W_f(a) \in \mathbb{Z}$ for all $a \in \mathbb{F}_2^n$), which means that a nonlinearity of $2^{n-1}-2^{n/2-1}$ can be reached for even $n$ only. Boolean functions in an even number of variables that achieve a nonlinearity of $2^{n-1}-2^{n/2-1}$ are called bent Boolean functions. 

The maximal possible nonlinearity for Boolean functions in odd dimensions lies between~\cite{carlet_2021}:
$2^{n-1}-2^{\frac{n-1}{2}}$ and $2\lfloor2^{n-2}-2^{\frac n 2-2}\rfloor$, where the first value is also called the quadratic bound. The highest nonlinearity for odd-sized Boolean functions is strictly larger than the quadratic bound for $n>7$~\cite{carlet_2021}.

\subsection{Monotone Boolean functions}
 
The function $f$ is called \emph{monotone increasing} if~\cite{carlet_2021}:
\begin{equation}
    u \preceq v \Longrightarrow f(u)\le f(v).
\label{eq:monotone}
\end{equation}
Equivalently, changing any input bit from $0$ to $1$ cannot change the output from
$1$ to $0$. 

\subsubsection{Nonlinearity of Monotone Boolean Functions.}
A first general result is that for every odd $n\ge 5$ and every monotone Boolean
function $f$ in $n$ variables~\cite{carlet2016cryptographic},
\begin{equation}
    nl_{f} \le 2^{n-1}-2^{(n-1)/2}.
\end{equation}

For even dimensions, it is shown that for every even $n\ge 10$ and every monotone
Boolean function $f$~\cite{carlet2016cryptographic},
\begin{equation}
   nl_{f}\le 2^{n-1}-2^{n/2}. 
\end{equation}
Thus, monotone Boolean functions are always significantly closer to affine functions
than bent functions are.

An even stronger bound, valid for every $n$, is obtained in terms of an auxiliary
quantity $M$~\cite{Carlet2017}:
\begin{equation}
    nl_{f}\le 2^{n-1}-\frac12\sqrt{M},
\end{equation}
where
\[
M=
\begin{cases}
\min(A,B,C), & \text{if } n \text{ is even},\\[1mm]
\min(B,C), & \text{if } n \text{ is odd},
\end{cases}
\]
with
\[
A
=
2^n
+
2\sum_{1\le j<n/4}\binom{n/2}{j}
\left(
2^{n/2}-2\sum_{i=0}^{j}\binom{n/2}{i}
\right)^2
+
(2^{n/2}-2)^2,
\]

\[
B
=
\min_{\substack{1\le k\le \frac{n}{2}\\ n+k\ \mathrm{even}}}
\left(
2^{n+k}
+
\sum_{j=\lfloor \frac{n+k}{4}\rfloor +1}^{\frac{n-k}{2}}
\binom{\frac{n-k}{2}}{j}
\left(
2^{\frac{n+k}{2}}
-
2\sum_{i=0}^{\frac{n+k}{2}-j}\binom{\frac{n+k}{2}}{i}
\right)^2
\right),
\]
and
\[
C=
\begin{cases}
\left(2\binom{n-1}{n/2}\right)^2, & \text{if } n \text{ is even},\\[2mm]
\left(2\binom{n-1}{(n-1)/2}\right)^2, & \text{if } n \text{ is odd}.
\end{cases}
\]

\subsubsection{Threshold and Majority Functions}

A central family of monotone Boolean functions is given by threshold functions~\cite{carlet2022complete}.
For positive integers $d\le n+1$, define
\[
T_{d,n}(x)=
\begin{cases}
0, & \text{if } w_H(x)<d,\\
1, & \text{if } w_H(x)\ge d,
\end{cases}
\qquad x\in\mathbb{F}_2^n.
\]
Thus, $T_{d,n}$ outputs $1$ exactly when the Hamming weight of the input is at least
$d$. These functions are symmetric and monotone increasing.

A special case of a threshold function is the majority function:
\[
\operatorname{MAJ}_n(x)=
\begin{cases}
0, & \text{if } w_H(x)\le \lfloor n/2\rfloor,\\
1, & \text{otherwise}.
\end{cases}
\]

\subsubsection{Exact Nonlinearity of Threshold Functions}

The nonlinearity of threshold functions is known exactly~\cite{carlet2022complete}. For every $n>0$ and
$1\le d\le n$,
\[
nl_{f}(T_{d,n})=
\begin{cases}
2^{n-1}-\binom{n-1}{(n-1)/2},
& \text{if } d=\dfrac{n+1}{2},\\[3mm]
\displaystyle\sum_{k=d}^{n}\binom{n}{k},
& \text{if } d>\dfrac{n+1}{2},\\[4mm]
\displaystyle\sum_{k=0}^{d-1}\binom{n}{k},
& \text{if } d<\dfrac{n+1}{2}.
\end{cases}
\]
Thus, the nonlinearity profile of threshold functions is symmetric with respect to the transformation $d\mapsto n-d+1$.

For odd $n$, the majority function is the unique central threshold function, and its
nonlinearity is
\[
nl_{f}
=
2^{n-1}-\binom{n-1}{(n-1)/2}.
\]
For even $n$, the majority function is
$\operatorname{MAJ}_n=T_{n/2+1,n}$,
and its nonlinearity equals
\[
nl_{f}
=
\sum_{k=n/2+1}^{n}\binom{n}{k}.
\]
By symmetry, this is also equal to the nonlinearity of the adjacent threshold
function $T_{n/2,n}$. We provide information about maximal known nonlinearity for balanced, imbalanced, and majority functions, as well as the upper bound for general Boolean functions and monotone functions in Table~\ref{tab:maj-thr-5-14}. Note that the bound for monotone functions results in non-integer values, while the nonlinearity values must be integers~\cite{Carlet2017}.
More information about Boolean functions can be found in, e.g.,~\cite{MacWilliams-Sloane,carlet_2021}.

\begin{table}[t]
\centering
\scriptsize
\caption{Max possible and known nonlinearities.}
\label{tab:maj-thr-5-14}
\renewcommand{\arraystretch}{1.15}
\setlength{\tabcolsep}{4pt}
\begin{tabular}{lcccccccccc}
\toprule
dimension & 5 & 6 & 7 & 8 & 9 & 10 & 11 & 12 & 13 & 14 \\\midrule
$nl_{f}$ (balanced) & 12 & 26 & 56 & 116 & 240 &492 &992 & 2010 &4036 &8120\\\midrule
$nl_{f}$ (imbalanced) & 12 & 28 & 56 & 120 & 242 & 496 &996 &2016 & 4040 &8128\\\midrule
$nl_{f}$ (majority) & 10 & 22 & 44 &93 &186 & 386 & 772 & 1586 & 3172 & 6476\\\midrule
$nl_{f}$ (bound) & 12 & 28 & 58 & 120 & 244 & 496 & 1000 & 2016 & 4050 & 8128\\\midrule
$nl_{f}$ (bound monotone) & 12 & 27 & 55.5 & 114.4 & 237.1 & 478.5 & 977.6 & 1975.1 & 3975.2 & 8013.1\\\bottomrule
\end{tabular}
\end{table}

\section{Related Work}
\label{sec:related}

Metaheuristics are widely used in the design of Boolean functions. Common application settings include cryptography~\cite{carlet_2021}, combinatorics~\cite{Rothaus}, and coding theory~\cite{KERDOCK1972182,MacWilliams-Sloane}. For a recent overview of metaheuristic approaches for the design of Boolean functions, we refer readers to~\cite{Djurasevic2023}.
To our knowledge, no prior work has considered the evolution of monotone Boolean functions.
However, many works consider evolving maximally nonlinear (bent) or balanced and highly nonlinear Boolean functions. 
We recall some of those works below.

In 1998, Millan et al. used heuristics to design cryptographically strong balanced Boolean functions~\cite{10.1007/BFb0054148}.
In 2003, Fuller et al. used evolutionary algorithms to evolve bent Boolean functions to 2003~\cite{FullerDM03}.
In 2005, Yang et al. used evolutionary algorithms and the trace representation of Boolean functions to evolve bent Boolean functions~\cite{cryptoeprint:2005/322}.

Picek and Jakobovic used genetic programming to evolve secondary algebraic constructions that are then used to construct bent Boolean functions~\cite{10.1145/2908812.2908915}. Hrbacek and Dvorak used Cartesian Genetic Programming to evolve bent Boolean functions~\cite{10.1007/978-3-319-10762-2_41}. Picek et al. considered the evolution of bent quaternary functions, i.e., a generalization of maximally nonlinear functions with a four-valued range~\cite{picek18a}. Mariot et al. used evolutionary strategies to evolve a secondary construction based on cellular automata for quadratic bent functions~\cite{MariotSLM22}.
Husa and Dobai used linear genetic programming to evolve bent Boolean functions~\cite{10.1145/3067695.3084220}.
Carlet et al. used several evolutionary algorithms and showed it is possible to evolve (anti)-self-dual bent functions for Boolean functions~\cite{Carlet2024}.
Yan et al. developed an improved genetic algorithm to construct weightwise (almost) perfectly balanced Boolean functions~\cite{10.1145/3579856.3590337}.
Carlet et al. provided a systematic evaluation of evolving highly nonlinear Boolean functions in odd dimensions~\cite{Carlet2025}.

\section{Methodology}
\label{sec:methodology}

\subsection{Solution Encodings}

We consider three encodings. Truth table encoding (TT) explores the full space and is maximally expressive, but unconstrained. 
Balanced truth table encoding (TTw) enforces balancedness directly, but may make monotonicity much harder to satisfy simultaneously. Finally, symbolic encoding used by genetic programming (GP) may better capture structural regularities of monotone functions through symbolic compositions.

\subsubsection{Truth table Encoding (TT).}

The most common way to represent a Boolean function is via its truth table (TT)~\cite{Djurasevic2023}, which is encoded as a bitstring. 
For a Boolean function with $n$ inputs, the truth table is a bitstring of length $2^n$.
The search algorithm explores the full space of $n$-variable Boolean functions, which has the size $2^{2^n}$. 

In this encoding, we use the simple bit mutation and the shuffle mutation. For crossover, we employ one-point and uniform crossover operators. Each time the evolutionary algorithm invokes a crossover or mutation operation, one of the previously described operators is randomly selected.

\subsubsection{Balanced Truth Table Encoding (TTw).}
The \textit{balanced} encoding also uses a truth table to represent a Boolean function.
However, the number of ones in the truth table is kept at the desired value ($2^n/2$) at all times, so that the corresponding function is always balanced.

This encoding is implemented with customized genetic operators that preserve this property, which includes initialization, mutation, and crossover.
The crossover and mutation operators used in this work are based on~\cite{manzoni20}, so that all genetic operators preserve the number of ones throughout the evolution. 
For the crossover, we use the balanced weighted crossover operator, and the mutation uses a two-bit inversion (which flips two random bits) and a mixing mutation, which shuffles the genes between two randomly chosen positions in the bitstring.

\subsubsection{Symbolic Encoding (GP).}
This encoding uses tree-based genetic programming (GP) to represent a Boolean function in its symbolic form.
A candidate solution is represented with a tree whose leaves correspond to the input variables $x_1,\ldots, x_n \in \F_2$. The internal nodes are Boolean operators that combine the inputs received from their children and forward their output to the respective parent nodes. 
We use the following function set: OR, XOR, AND, IF, and the NOT function that takes a single argument. 
The function IF takes three arguments and returns the second one if the first one evaluates to true, and the third one otherwise. 
This function set is common in previous applications and is based on our tuning results.
The output of the root node is the output value of the Boolean function. The truth table of the function $f: \F_2^n \to \F_2$ is determined by evaluating the tree over all possible $2^n$ assignments of the inputs at the leaves. 

For the symbolic encoding, the genetic operators used in our experiments are simple tree crossover, uniform crossover, size fair, one-point, and context preserving crossover~\cite{poli08:fieldguide} (selected at random), and subtree mutation.
Multiple genetic operators were used based on previous results indicating better convergence when using a diverse set of operators.

\subsection{Fitness Functions}
\label{subsec:fit}

\subsubsection{Non-monotone Penalty}

In all experimental configurations, we use the non-monotone penalty to drive the search towards monotone increasing Boolean functions.
One way to enforce this property is to penalize the function for every case where the monotone requirement (see Eq.~\eqref{eq:monotone}) is violated.
This is calculated with the following:
\begin{itemize}
    \item iterate over every input $u$ for which $f(u) = 1$;
    \item for each such input $u$, iterate over every $v$ obtained by taking each input bit equal to 0 and changing it to 1;
    \item increase penalty value $penalty$ by 1 for every case in which $f(v) = 0$.
\end{itemize}
If the $penalty$ is zero, meaning the function is monotone, then we add the desired property to optimize in the fitness function.

\subsubsection{Fitness for Balanced Functions}

In the first scenario, we search for balanced, highly nonlinear monotone functions.
All fitness functions include a non-monotone penalty term, $penalty$, that counts the occurrences that do not satisfy the monotone property.
This term is added with a negative sign to act as a penalty in maximization.
The delta function $\delta_{penalty, 0}$ assumes the value one when $penalty = 0$ and is zero otherwise. 

Furthermore, to obtain only balanced monotone functions, we include the balancedness penalty $BAL$, which is defined as the difference up to the balancedness (i.e., the number of bits to be changed to make the function balanced). 
This difference is also subtracted to facilitate maximization.
The delta function $\delta_{BAL, 0}$ assumes the value one when $BAL = 0$ and is zero otherwise. 
Only if the function is monotone and balanced, i.e., if \textit{both} $BAL$ and $penalty$ equal zero, the nonlinearity value ($nl_f$) is added, and the fitness is maximized.

However, rather than considering only the extreme values of the Walsh–Hadamard spectrum, we also observe the number of times the maximum absolute value occurs in the spectrum, denoted by $\#max\_vals$.
Since higher nonlinearity corresponds to a \textit{lower} maximal absolute value, we aim for as few occurrences of the maximal value as possible to reach the next nonlinearity value; therefore, we include the inverse percentage of these values as a non-integer part added to the nonlinearity value.
Finally, the composite fitness function is defined as:
\begin{equation}
\label{eq:fit}
fitness_{BAL} = -penalty -BAL + \delta_{penalty, 0} \cdot \delta_{BAL,0} \cdot \left( nl_{f} + \frac{2^n - \#max\_vals}{2^n} \right).
\end{equation}

\subsubsection{Fitness for Imbalanced Functions}

For the imbalanced monotone functions scenario, we can simply omit the balancedness penalty, thus arriving at the following fitness function: 
\begin{equation}
\label{eq:fit1}
fitness_{IMB} = -penalty + \delta_{penalty, 0} \cdot \left( nl_{f} + \frac{2^n - \#max\_vals}{2^n} \right).
\end{equation}

\begin{figure}[h]
    \centering

\begin{subfigure}[b]{0.49\textwidth}
\includegraphics[width=1.\linewidth]{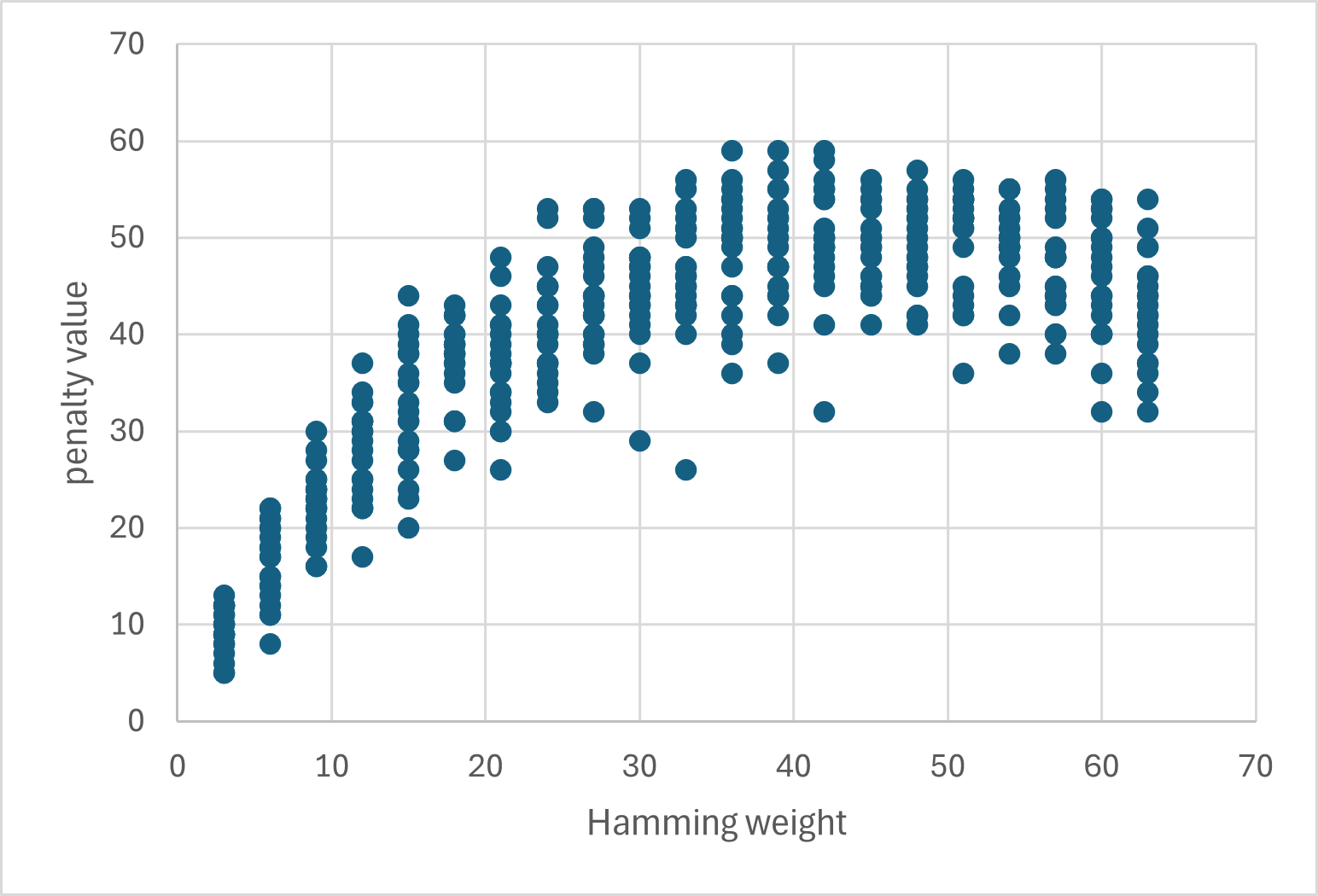} 
\caption{Penalty for $fitness_1$: count non-monotone occurences}
\label{fig:penalty1}
\end{subfigure}
\hfill
\begin{subfigure}[b]{0.49\textwidth}
\includegraphics[width=1.\linewidth]{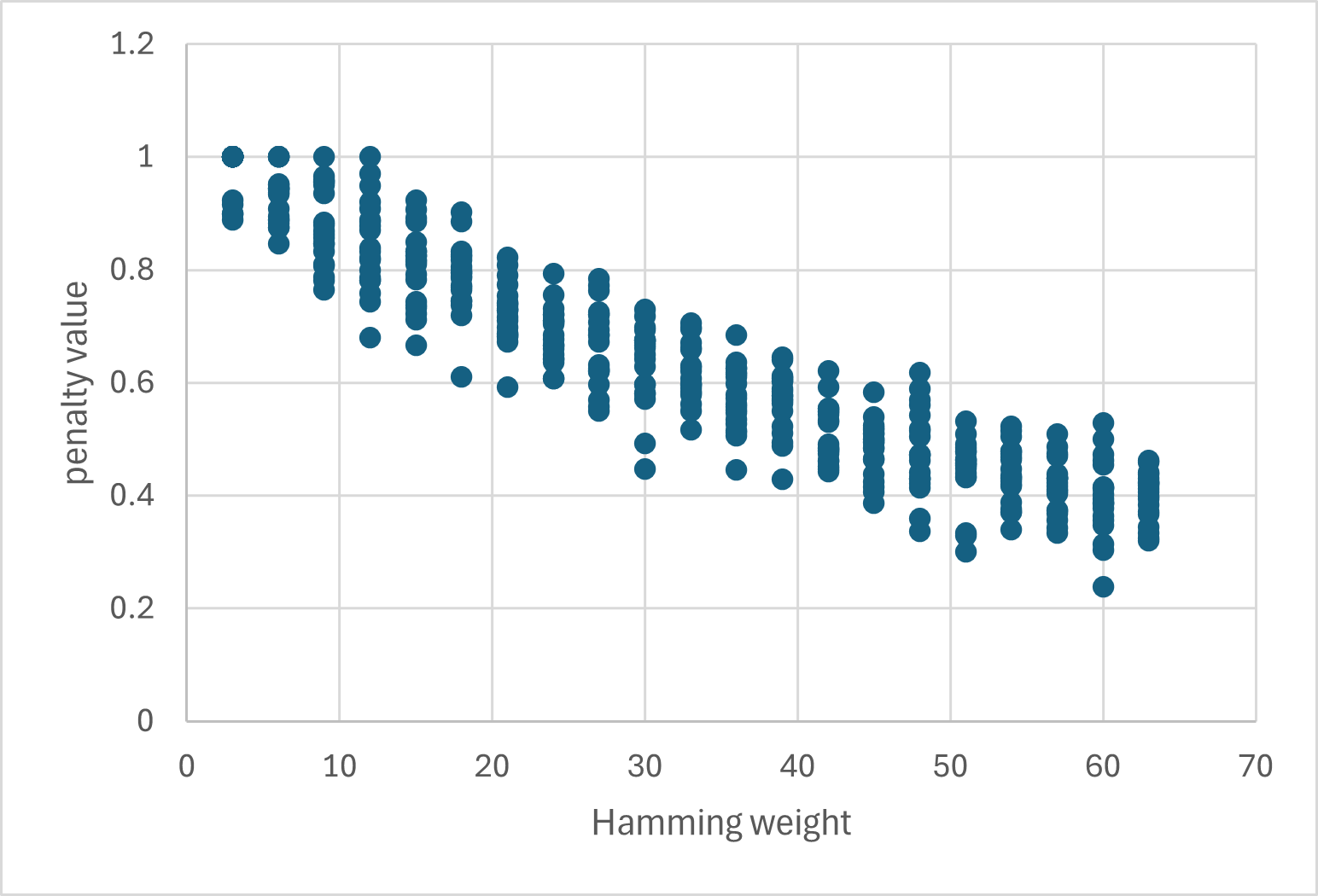}
\caption{Penalty for $fitness_2$: normalized by maximum penalty}
\label{fig:penalty2}
\end{subfigure}    
\hfill
\begin{subfigure}[b]{0.49\textwidth}
\includegraphics[width=1.\linewidth]{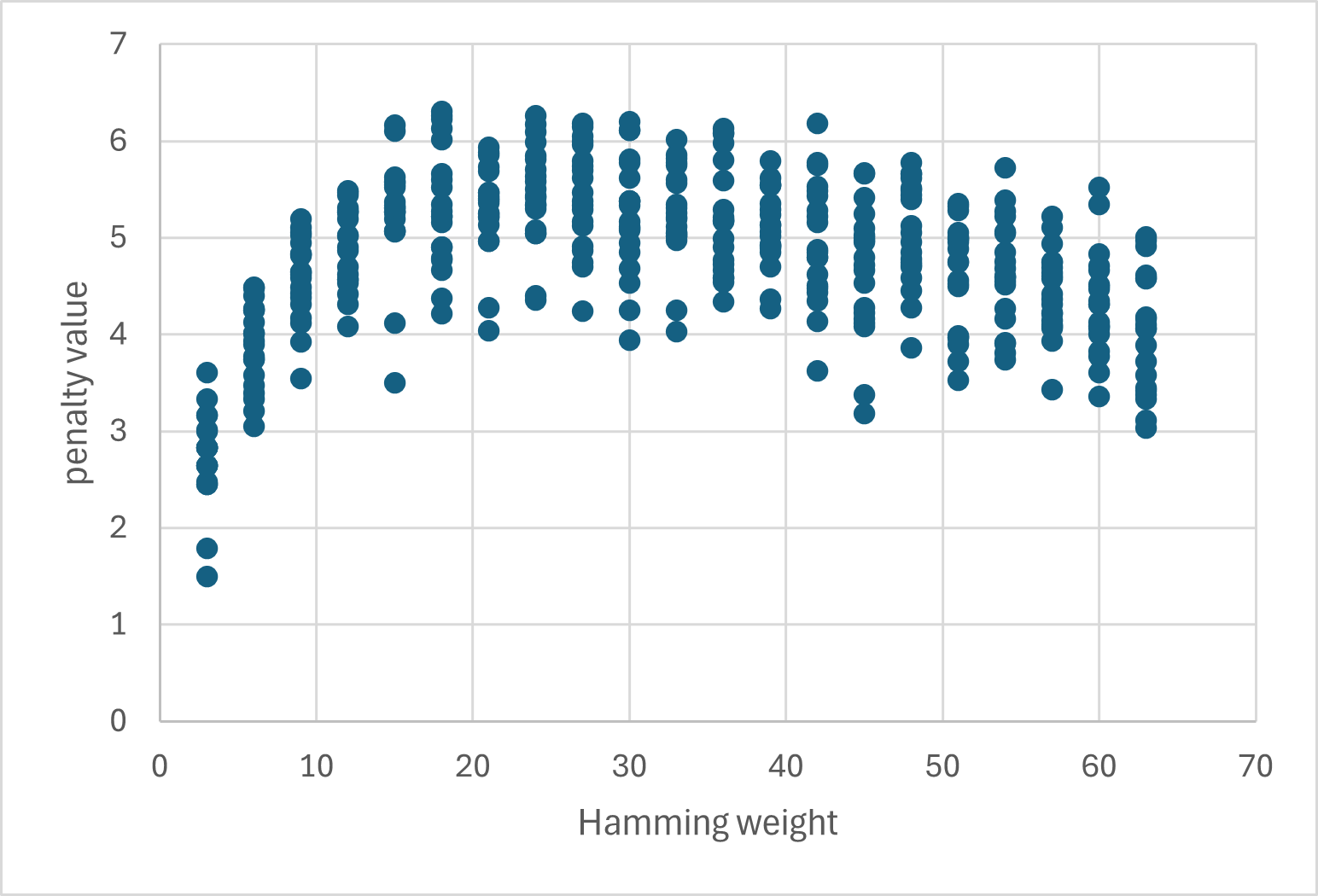}
\caption{Penalty for $fitness_3$: normalized by square of max. penalty}
\label{fig:penalty3}
\end{subfigure}    

    \caption{Non-monotone penalty variants.}
    \label{fig:penalty}
\end{figure}

However, when calculating the non-monotone penalty, it can be observed that Boolean functions with a lower Hamming weight will, on average, receive a lower penalty: since we count only cases where $f(u)$ changes from 1 to 0, functions having fewer ones in the truth table will have a lower maximum possible penalty.
This is evident from Figure~\ref{fig:penalty1}, where we sample random Boolean functions (of 6 variables) with different weights and record the obtained penalties; it is obvious that this penalty will favor functions with a lower Hamming weight.
To achieve a more uniform penalty pressure over all weight values, we experiment with two additional variants; $penalty_2$ takes the original count $penalty$ and divides it with the number of total \textit{possible} occurences of non-monotone violations, which is equal to the number of all truth table entries $u$ where $f(u) = 1$ and all possible bit changes from 0 to 1 in input $u$.
Finally, $penalty_3$ is divided by the square of the maximum possible penalty, since it was observed that $penalty_2$ introduces an inverted weight bias (Figure~\ref{fig:penalty}).
Therefore, the experiments for imbalanced monotone functions include three fitness functions, denoted $fit_1$, $fit_2$, and $fit_3$, which simply correspond to the penalty variant used in Eq.~\eqref{eq:fit1}.
Note that different penalty measures incur different scales, but this is irrelevant in cases where selection is not proportional.

The three penalty variants are motivated by the dependence of monotonicity violations on the Hamming weight of the candidate function. The raw penalty in \textit{fit1} counts all local violations directly and therefore favors functions with fewer ones in their truth table, since such functions have fewer opportunities to violate monotonicity. To reduce this bias, \textit{fit2} normalizes the count by the maximum possible number of violations for the given support size. As shown in Figure~\ref{fig:penalty2}, this correction is too strong and introduces the opposite bias, favoring functions with larger Hamming weight. 
We then introduce \textit{fit3}, which uses a milder normalization by the square of the maximum possible penalty. This yields a more uniform penalty landscape across Hamming weights and acts as a compromise between the two extremes.

\subsection{Algorithms and Parameters}
\label{sec:settings}

We use the same evolutionary algorithm for all encodings: a steady-state selection with a 3-tournament elimination operator. 
In each iteration, three individuals from the population are chosen at random, and the worst one in terms of fitness value is eliminated. 
The two remaining individuals are used with the crossover operator to generate a child individual, which then undergoes mutation with individual mutation probability $p_{mut} = 0.5$. The mutated child replaces the eliminated individual in the population.
All algorithms use the same stopping criterion of $10^6$ evaluations, and all experiments are executed in 30 runs.
The implementation was made using the ECF framework\footnote{http://ecf.zemris.fer.hr/}~\cite{ECF} and the code is available upon request.

\section{Experimental Results}
\label{sec:results}

\begin{table}
\caption{Best obtained results (nonlinearity) for all considered sizes.}
\renewcommand{\arraystretch}{1.25}
\setlength{\tabcolsep}{4pt}
\begin{center}
\label{tab:best}
\begin{tabular}
{@{}lcccccccccc@{}}
\toprule
              & 5  & 6  & 7  & 8  & 9   & 10  & 11  & 12   & 13   & 14   \\ \midrule
bal: TT       & 10 & 22 & \textbf{46} & \textbf{94} & \textbf{192} & \textbf{388} & \textbf{786} & \textbf{1583} & \textbf{2835} & -    \\
bal: TTw      & 10 & 22 & 44 & 70 & 34  & 22  & 18  & 14   & 12   & 6    \\
bal: GP       & 10 & 22 & 44 & 84 & 176 & 336 & 704 & 1281 & 2721 & \textbf{5441} \\ \midrule
imb: TT, fit1 & 11 & 23 & \textbf{47} & 96 & 195 & 396 & 797 & 1612 & 3173 & -    \\
imb: TT, fit2 & 11 & 23 & \textbf{47} & 96 & 195 & 396 & 800 & 1596 & 1156 & 498  \\
imb: TT, fit3 & 11 & 23 & \textbf{47} & \textbf{97} & 195 & 397 & 799 & 1610 & 2462 & -    \\
imb: GP, fit1 & 11 & 23 & 46 & \textbf{97} & 196 & 396 & \textbf{802} & 1581 & 3072 & 6651 \\
imb: GP, fit2 & 11 & 23 & 46 & 95 & 196 & \textbf{398} & 796 & 1619 & \textbf{3401} & \textbf{6750} \\
imb: GP, fit3 & 11 & 23 & 46 & 96 & \textbf{198} & 395 & 767 & \textbf{1642} & 3293 & 6659 \\ \bottomrule
\end{tabular}
\end{center}
\end{table}

In the experiments, we carry out experiments in two scenarios: optimization of balanced functions, with a single fitness function, and imbalanced Boolean functions with three fitness variants.
Experiments are made using three encodings (TT, TTw, and GP) and problem sizes from 5 to 14 variables.

As a concise overview, Table~\ref{tab:best} shows only the best obtained nonlinearity values for every combination of size, encoding, and fitness function.
The dash as the table entry denotes a configuration in which not a single monotone function was found in 30 runs.
From the values in the table, we can see that the evolutionary approach is effective at discovering monotone functions with nonlinearity substantially above the majority baseline. 
Indeed, for every tested size, we find monotone functions with higher nonlinearity than that of majority functions. 
This comparison is important because the majority functions are the most classical monotone benchmark, yet they substantially underestimate the nonlinear behavior achievable within the monotone class. In this sense, evolutionary search does not merely rediscover known constructions, but explores regions of the monotone-function space that are not captured by standard threshold-based examples.

Naturally, the obtained values still remain below the known upper bound for monotone functions (except for $n=5$), but the extent to which explicit monotone constructions can approach that bound remains open. 
At the same time, the remaining gap to the best known upper bound for monotone functions should be interpreted with caution: it may reflect limitations of the search process, but it may also indicate that the current upper bound is not tight for explicit constructions in the tested dimensions.
For sizes $n=5, 6$, all approaches work equally well, showcasing that the problem is still easy.

\begin{figure}
     \centering
     \begin{subfigure}[b]{0.45\textwidth}
         \centering
\includegraphics[trim=6cm 6cm 0cm 0cm, width=\textwidth]{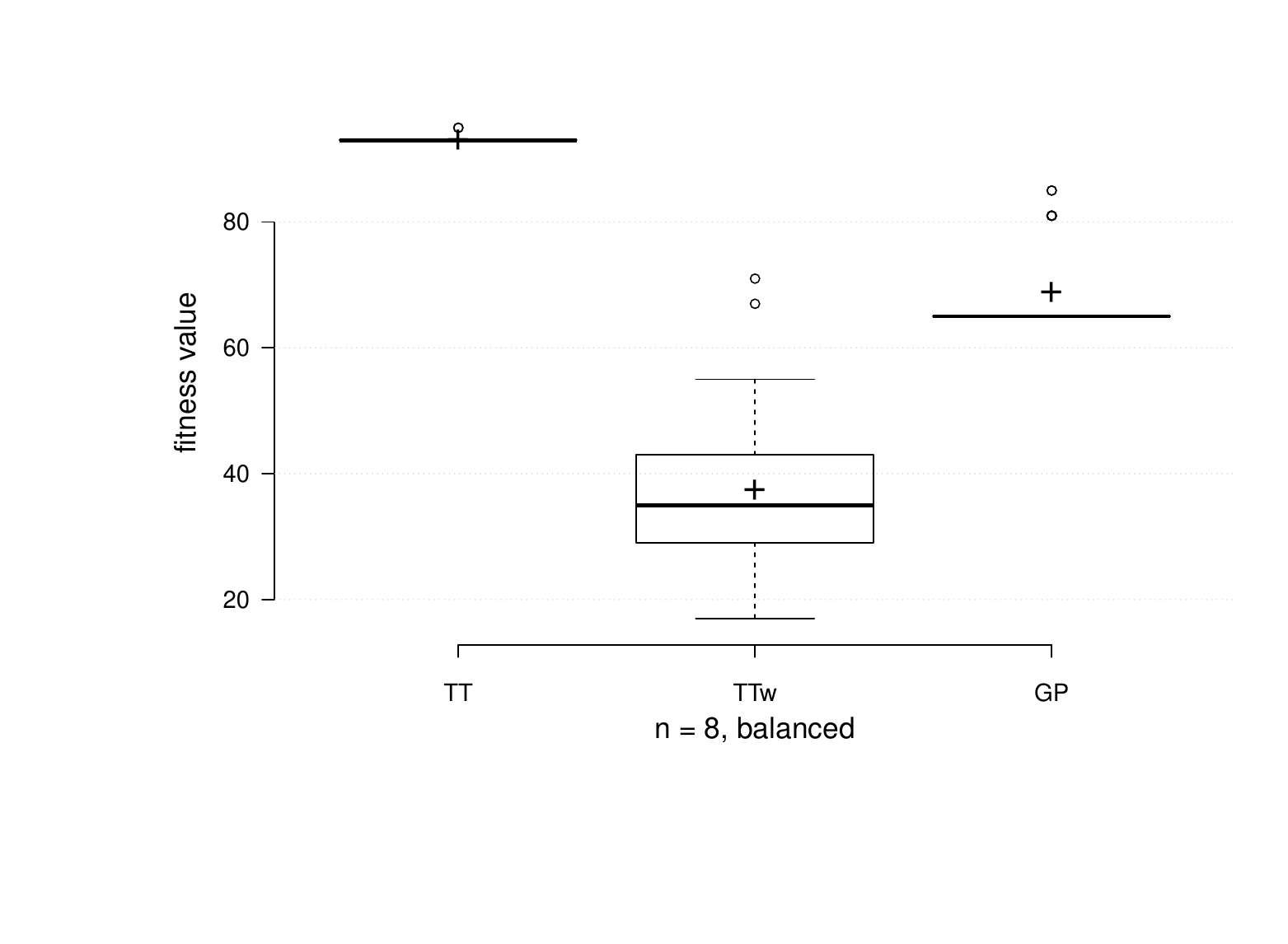}
\label{8bal}
     \end{subfigure}
     \hfill
     \begin{subfigure}[b]{0.45\textwidth}
         \centering
\includegraphics[trim=6cm 6cm 0cm 0cm, width=\textwidth]{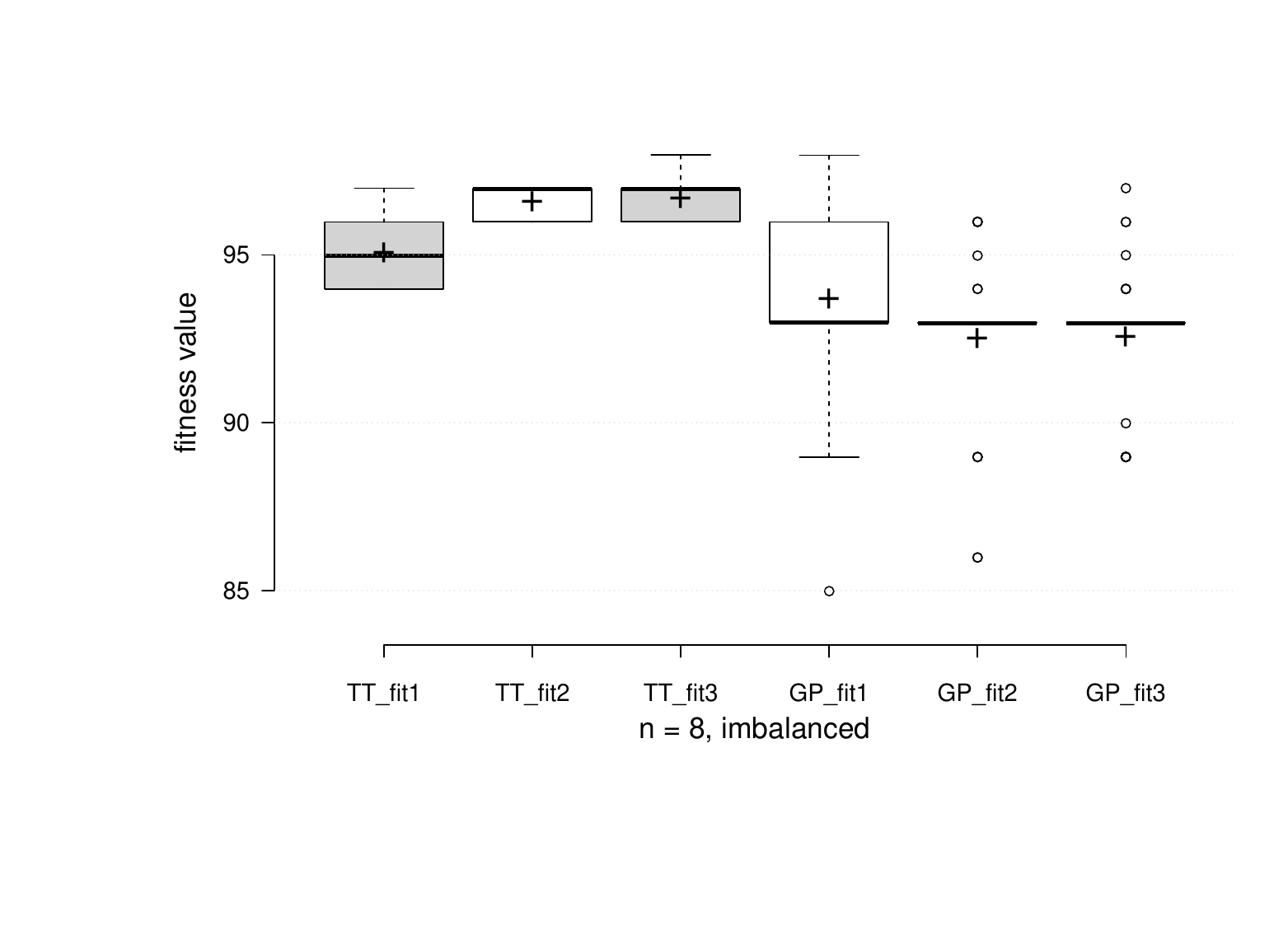}
\label{8imb}
     \end{subfigure}
        \caption{Results for $n=8$}
        \label{fig:8}
\end{figure}

\begin{figure}
     \centering
     \begin{subfigure}[b]{0.45\textwidth}
         \centering
\includegraphics[trim=6cm 6cm 0cm 0cm, width=\textwidth]{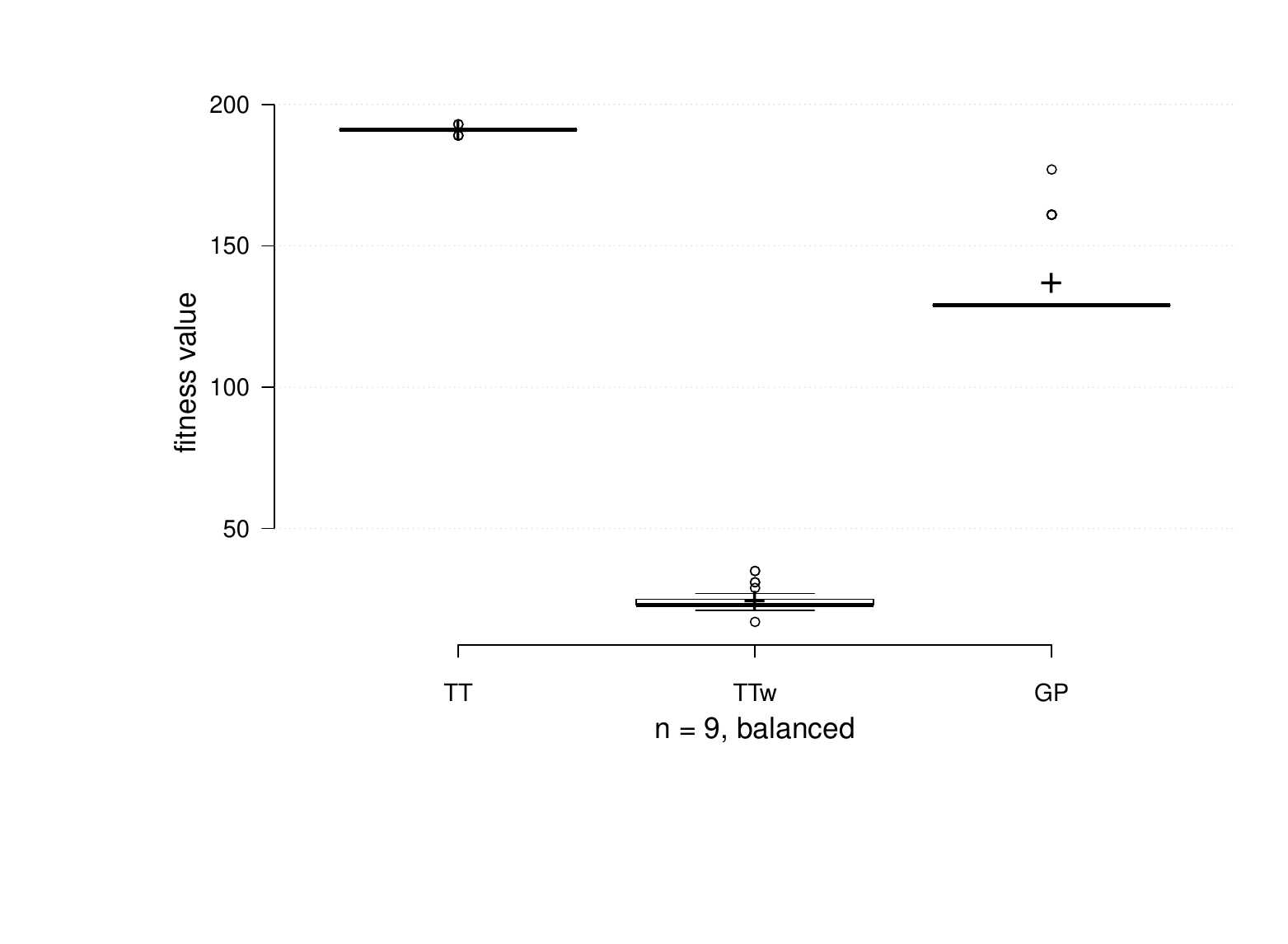}
\label{9bal}
     \end{subfigure}
     \hfill
     \begin{subfigure}[b]{0.45\textwidth}
         \centering
\includegraphics[trim=6cm 6cm 0cm 0cm, width=\textwidth]{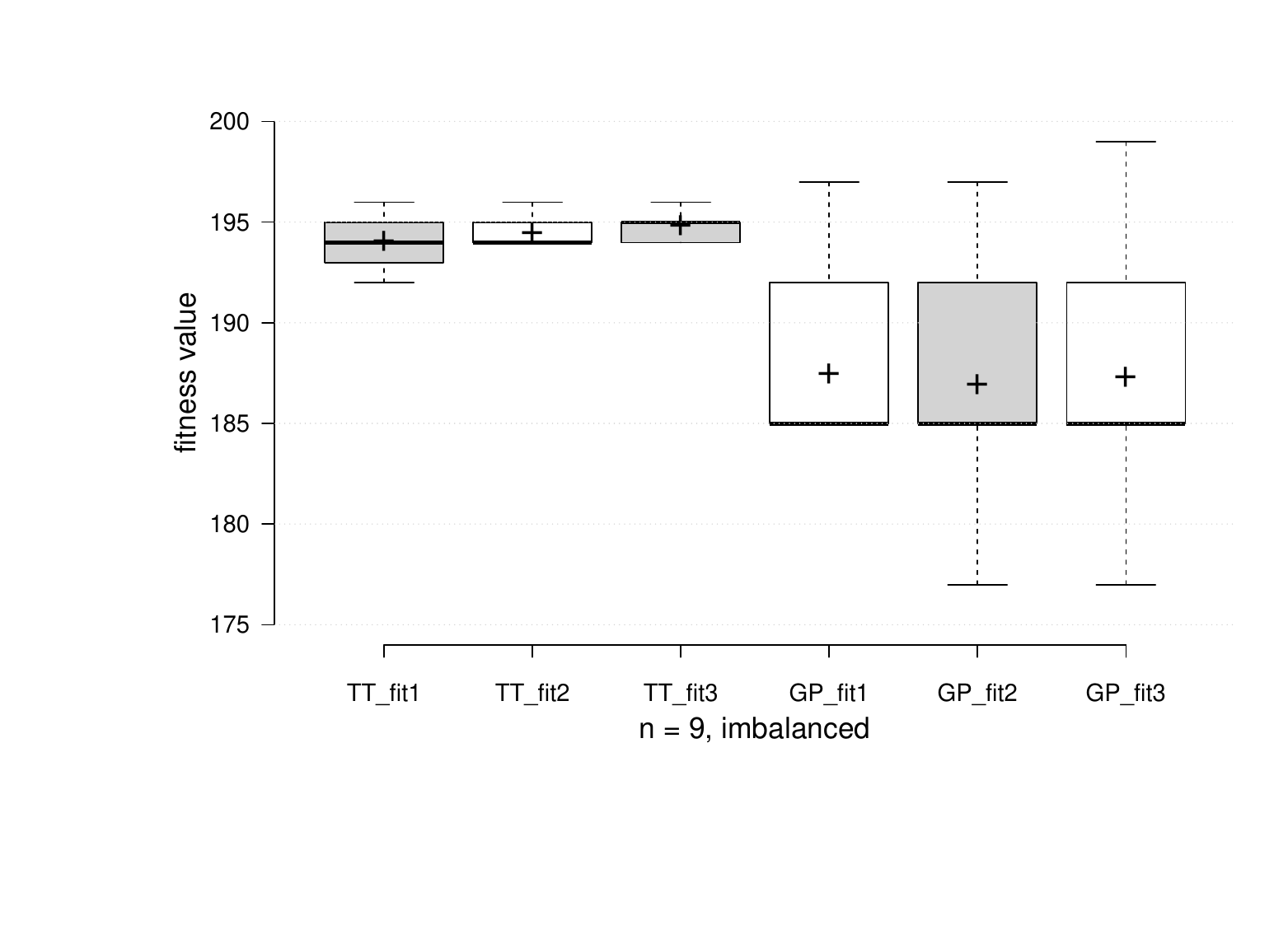}
\label{9imb}
     \end{subfigure}
        \caption{Results for $n=9$}
        \label{fig:9}
\end{figure}

\begin{figure}
     \centering
     \begin{subfigure}[b]{0.45\textwidth}
         \centering
\includegraphics[trim=6cm 6cm 0cm 0cm, width=\textwidth]{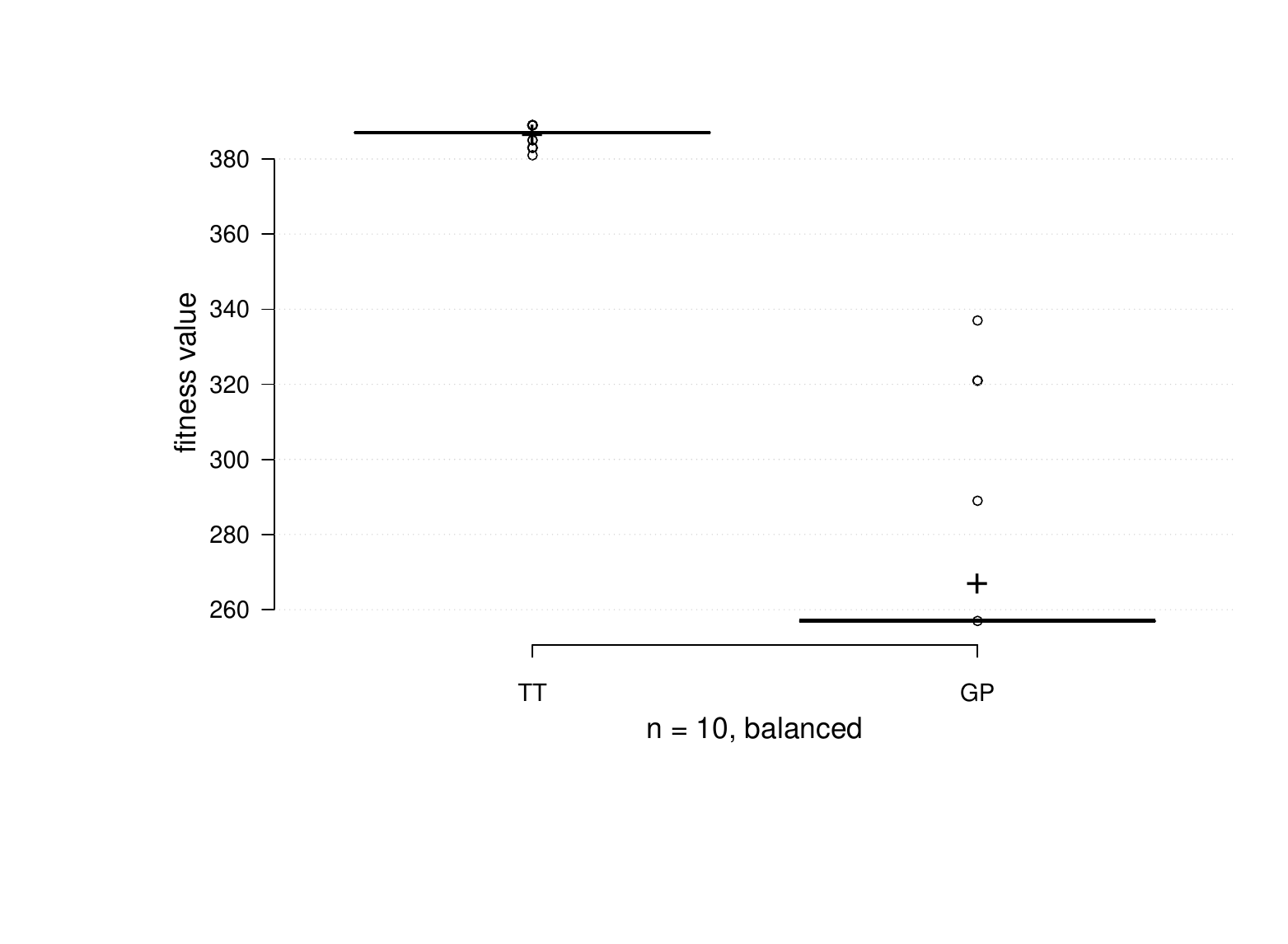}
\label{10bal}
     \end{subfigure}
     \hfill
     \begin{subfigure}[b]{0.45\textwidth}
         \centering
\includegraphics[trim=6cm 6cm 0cm 0cm, width=\textwidth]{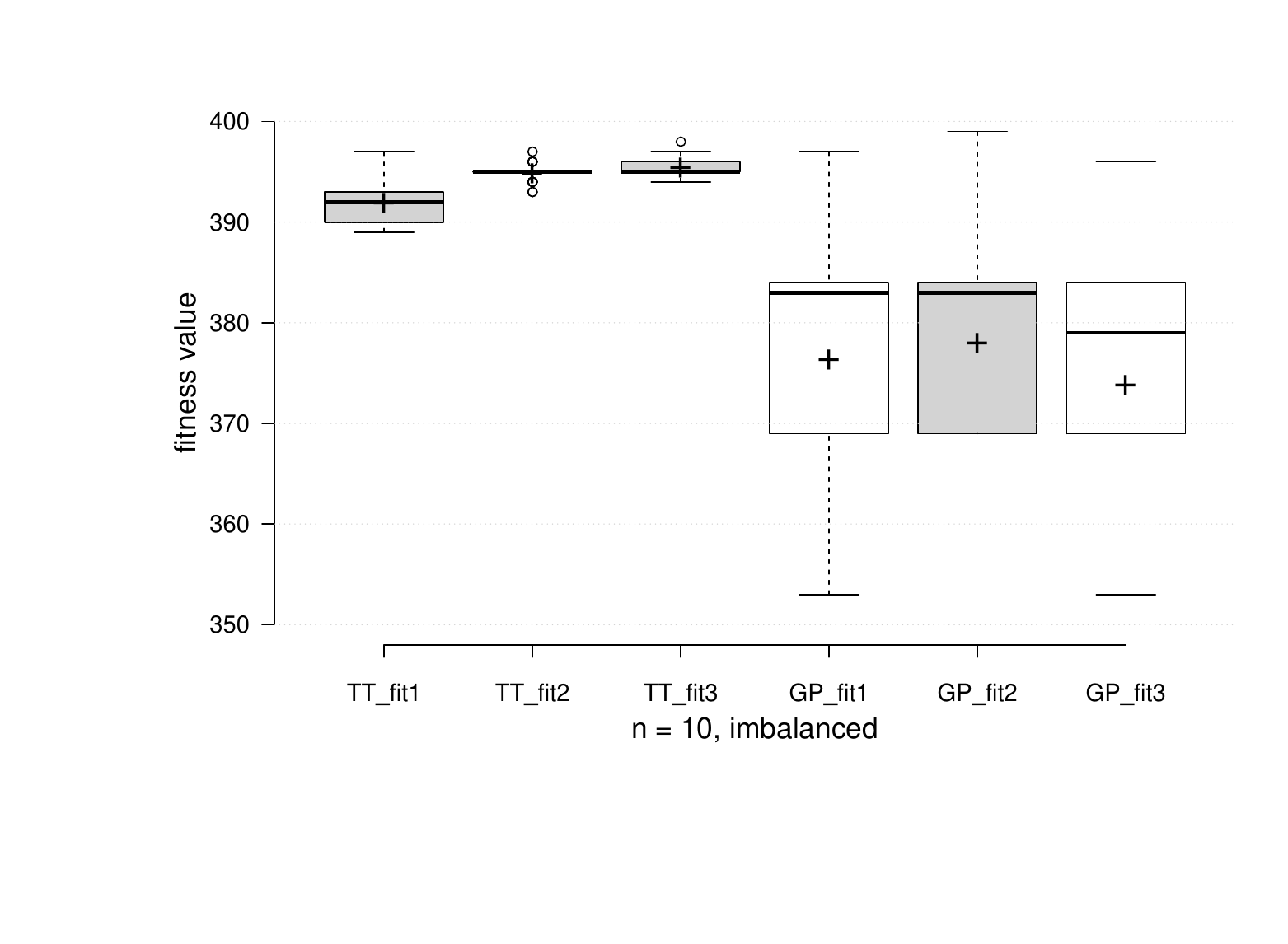}
\label{10imb}
     \end{subfigure}
        \caption{Results for $n=10$}
        \label{fig:10}
\end{figure}

\begin{figure}
     \centering
     \begin{subfigure}[b]{0.45\textwidth}
         \centering
\includegraphics[trim=6cm 6cm 0cm 0cm, width=\textwidth]{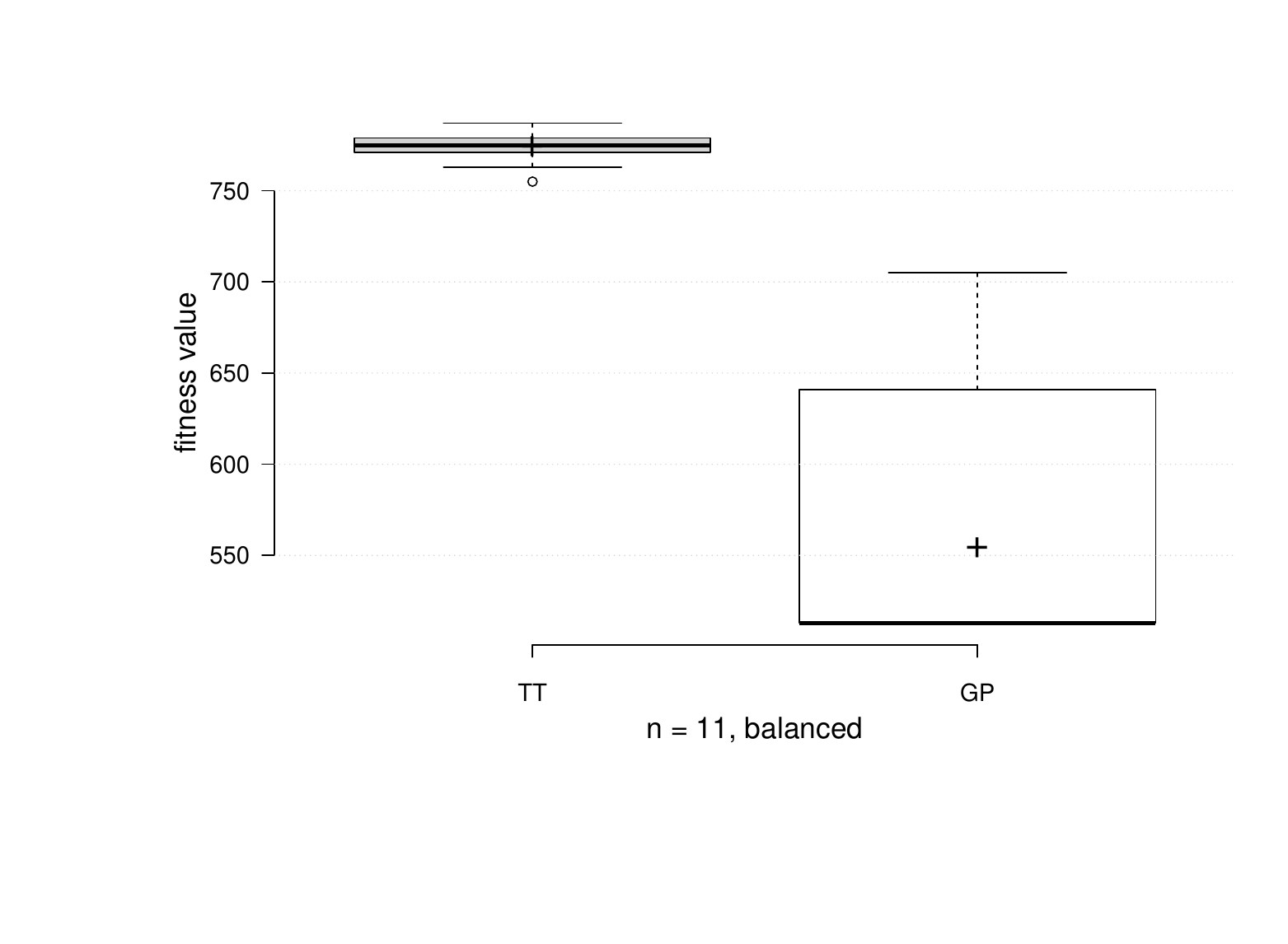}
\label{11bal}
     \end{subfigure}
     \hfill
     \begin{subfigure}[b]{0.45\textwidth}
         \centering
\includegraphics[trim=6cm 6cm 0cm 0cm, width=\textwidth]{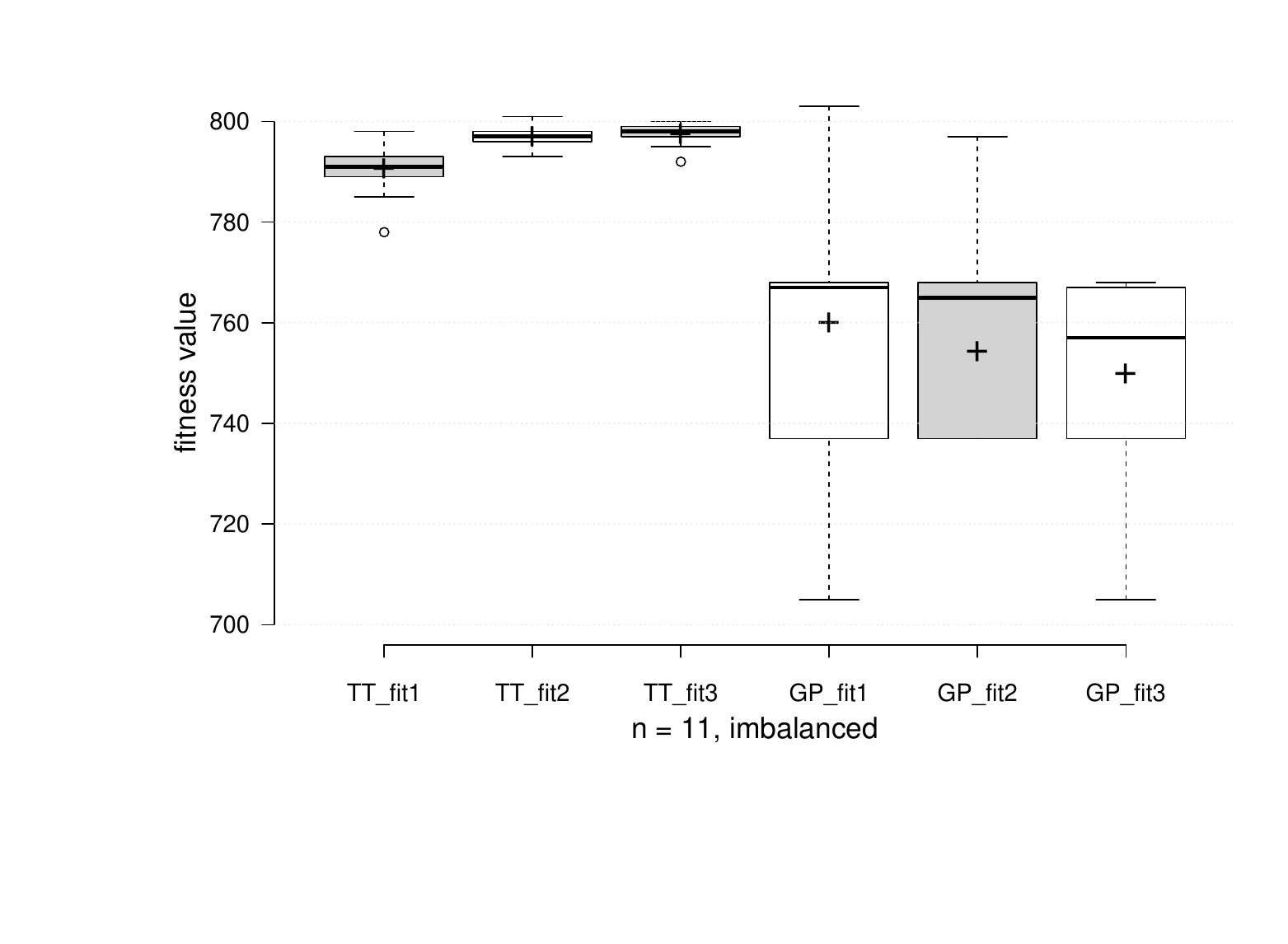}
\label{11imb}
     \end{subfigure}
        \caption{Results for $n=11$}
        \label{fig:11}
\end{figure}

\begin{figure}
     \centering
     \begin{subfigure}[b]{0.45\textwidth}
         \centering
\includegraphics[trim=6cm 6cm 0cm 0cm, width=\textwidth]{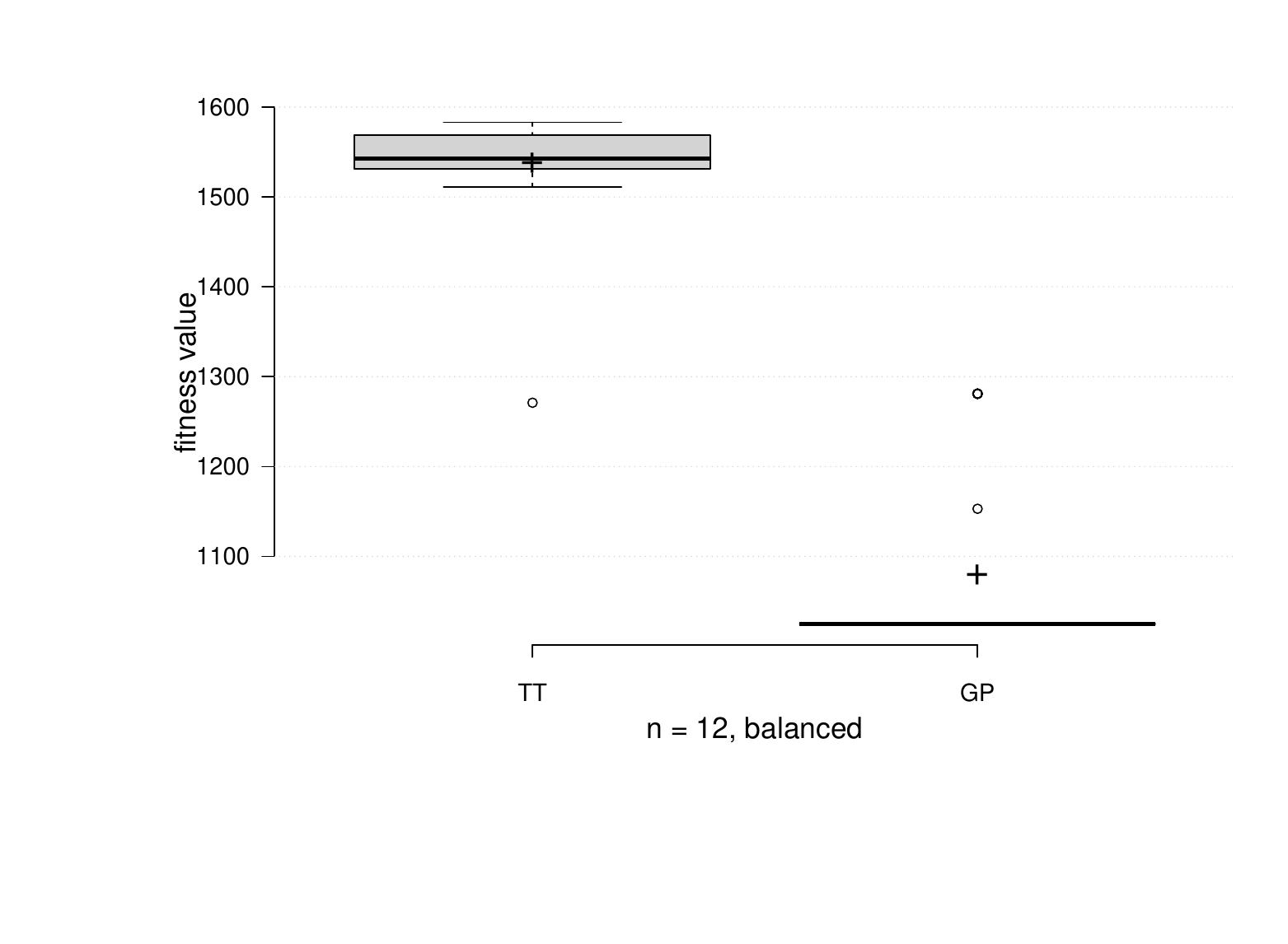}
\label{12bal}
     \end{subfigure}
     \hfill
     \begin{subfigure}[b]{0.45\textwidth}
         \centering
\includegraphics[trim=6cm 6cm 0cm 0cm, width=\textwidth]{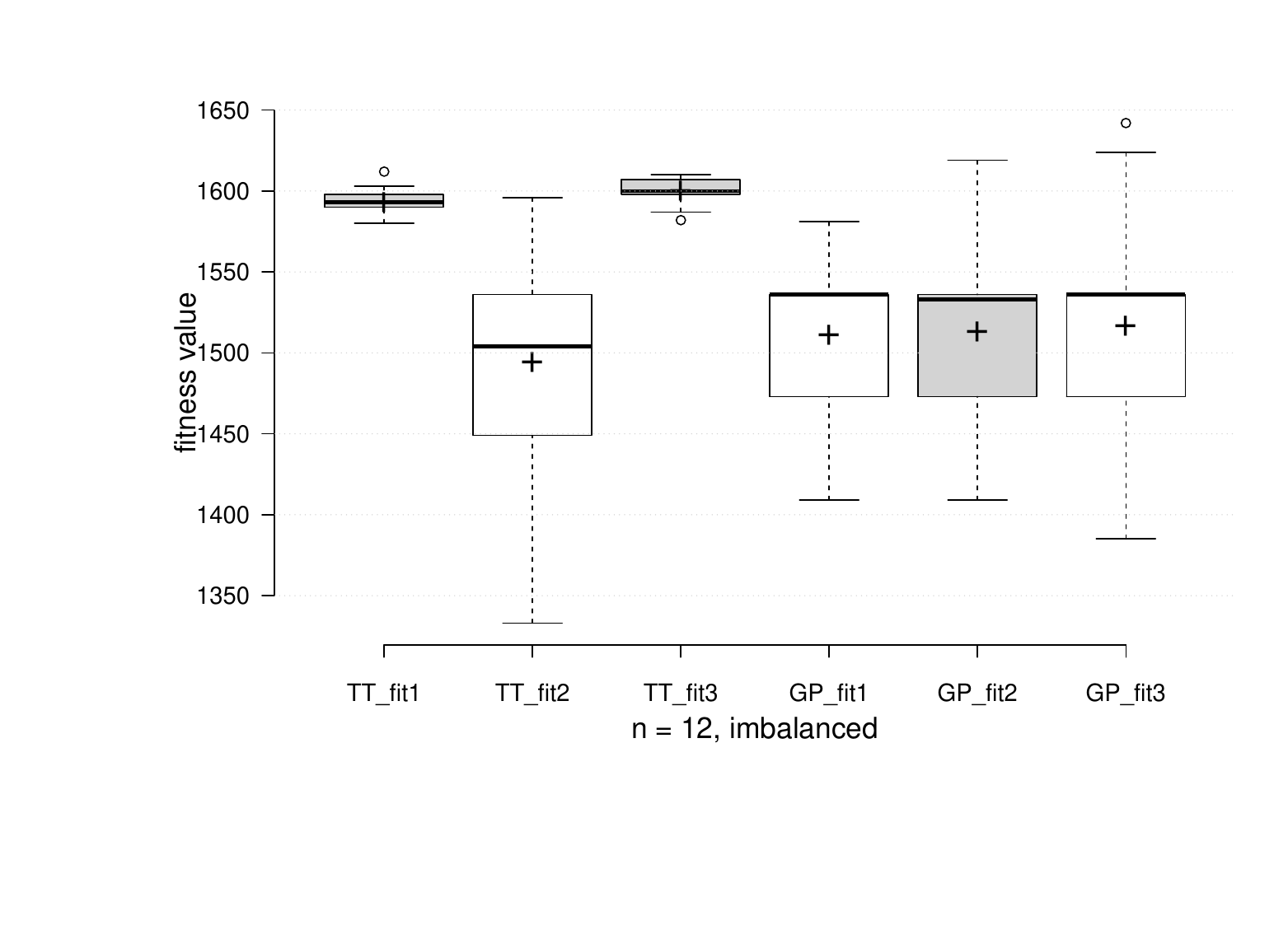}
\label{12imb}
     \end{subfigure}
        \caption{Results for $n=12$}
        \label{fig:12}
\end{figure}

\begin{figure}
     \centering
     \begin{subfigure}[b]{0.45\textwidth}
         \centering
\includegraphics[trim=6cm 6cm 0cm 0cm, width=\textwidth]{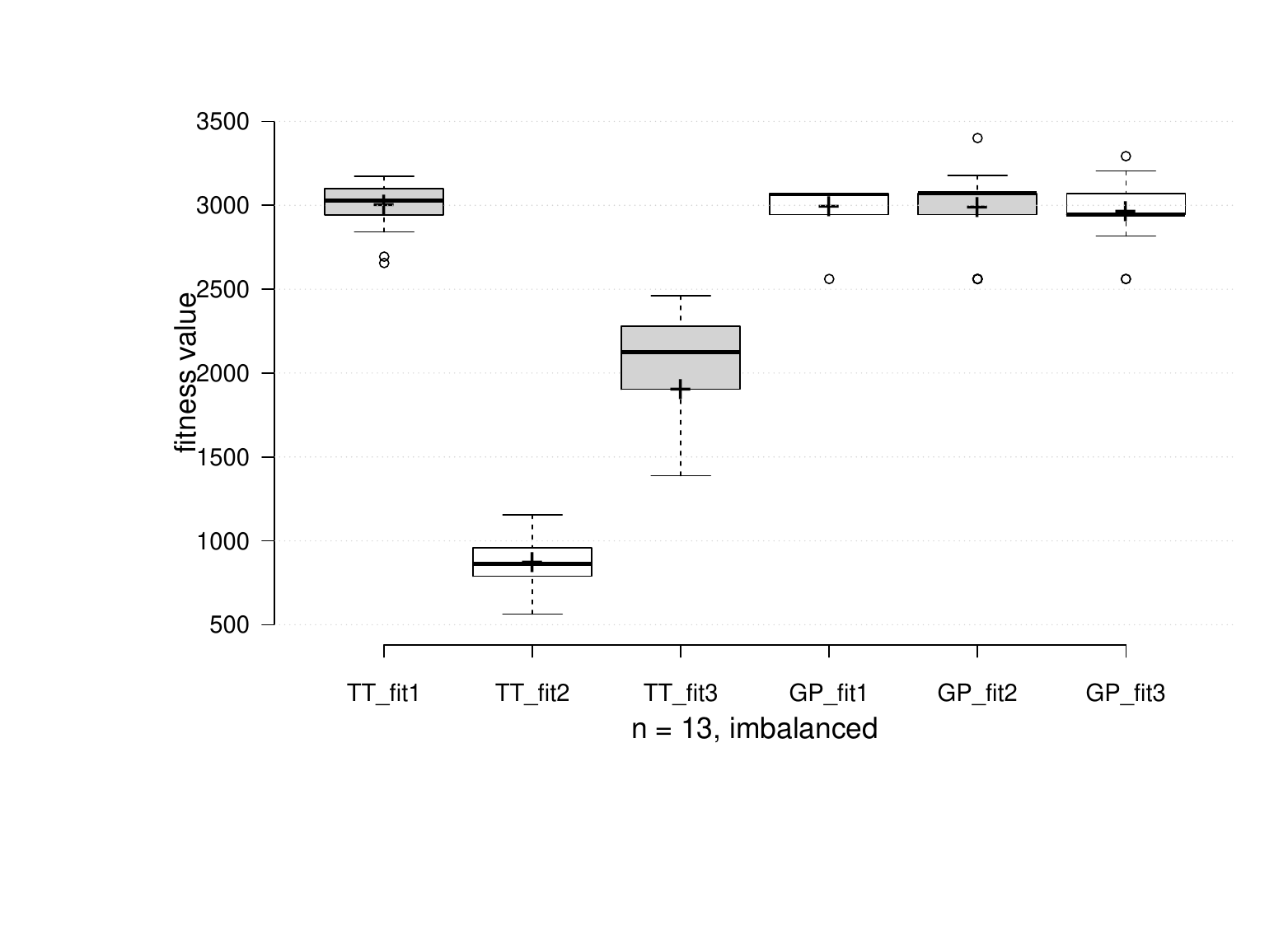}
\label{13imb}
     \end{subfigure}
     \hfill
     \begin{subfigure}[b]{0.45\textwidth}
         \centering
\includegraphics[trim=6cm 6cm 0cm 0cm, width=\textwidth]{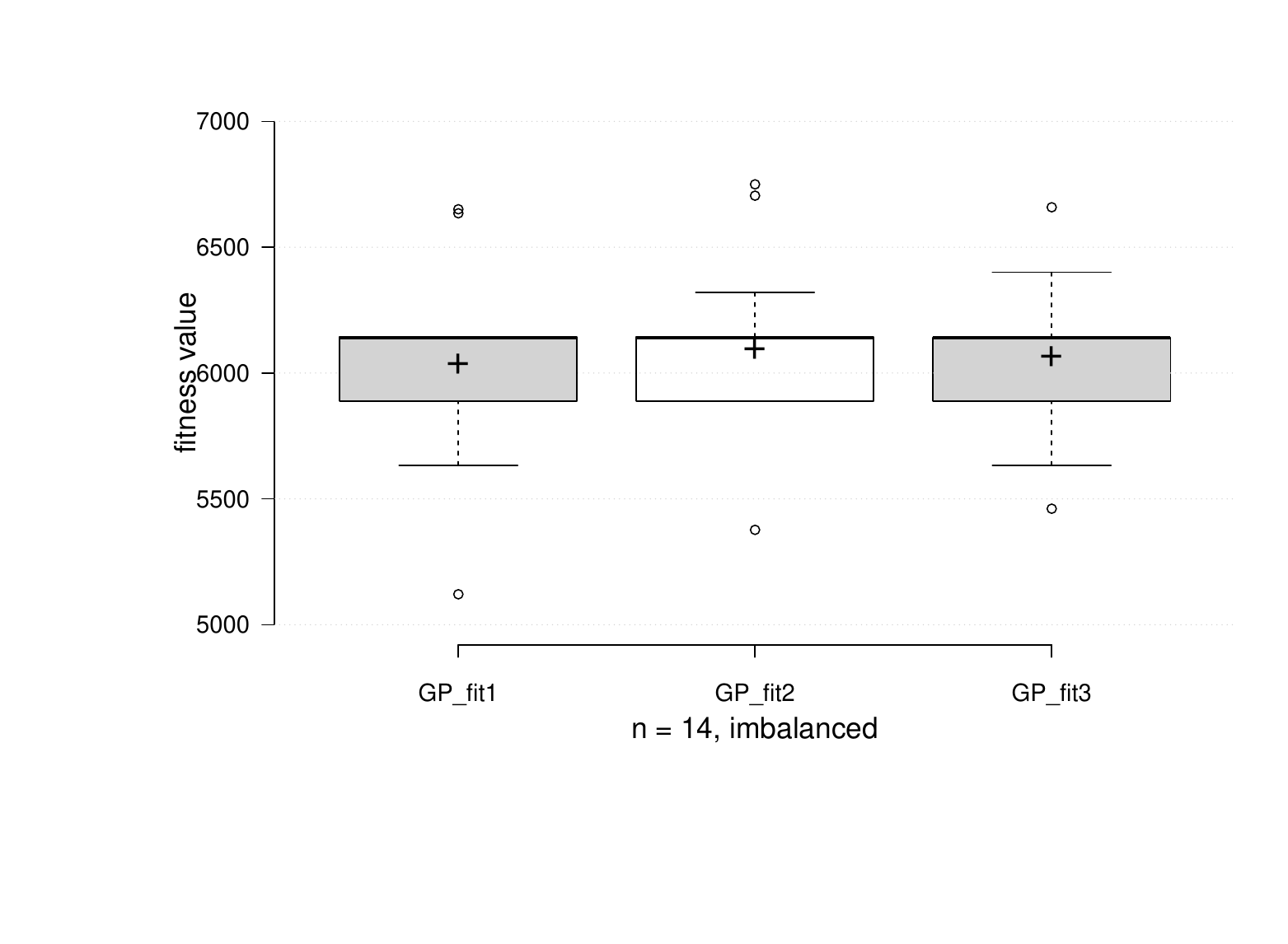}
\label{14imb}
     \end{subfigure}
        \caption{Results for $n=13$ and $n=14$, imbalanced functions}
        \label{fig:13}
\end{figure}

The results for individual problem sizes are presented in Figures~\ref{fig:8} to~\ref{fig:13}.
The results for the balanced scenario show that the balanced encoding (TTw) with enforced weight performs worse than TT for $n \le 7$, and worse overall for all larger $n$.
This can also be observed in Table~\ref{tab:best}, since even the best TTw results are much smaller than the ones from other encodings.
For that reason, we omit the TTw values in figures depicting balanced results for a larger number of variables.
The balanced scenario is visibly harder than the imbalanced one, which is natural because the search must simultaneously satisfy two strong constraints: monotonicity and exact balance. This helps explain both the rapid deterioration of the TTw encoding and the fact that, in larger dimensions, even the unrestricted TT representation struggles to locate feasible balanced monotone solutions.

As for the symbolic encoding (GP) compared to TT, we can see that it is generally worse than TT, but it starts to perform competitively for sizes 13 and 14.
Indeed, already for $n=13$, the balanced TT cannot find a balanced monotone function in most of the runs (which end with a negative fitness value); only in a single run (out of 30), the best nonlinearity value of 2835 is found with TT, most likely by pure chance.
For this reason, the plot for the balanced case in 13 variables is not shown (since the majority of TT results are negative).

Furthermore, in $n=14$ variables, the TT encoding does not succeed in finding a single monotone Boolean function, whether balanced or not; therefore, the plot for this case (Figure~\ref{fig:13}) shows only GP results.
This difference between TT and GP suggests that representation bias becomes increasingly important as the dimension grows: while TT is highly expressive, GP appears better able to exploit structural regularities of monotone functions once the feasible region becomes too sparse to be reached reliably by direct truth table search.

Throughout the experiments, fitness functions \textit{fit2} and \textit{fit3} offer a slight advantage over \textit{fit1} in the imbalanced setting; however, this effect is visible mainly for the TT encoding and mostly for sizes below $n=13$. This suggests that correcting the Hamming weight bias in the penalty is useful when the search is performed directly in truth table space, where local changes strongly affect support size. In contrast, for GP, the structural bias induced by the representation appears to dominate the influence of the penalty normalization, so the three fitness variants behave much more similarly.

\section{Conclusions and Future Work}
\label{sec:conclusions}

In this work, we studied the problem of evolving monotone Boolean functions with high nonlinearity. Unlike the classical setting of unrestricted Boolean-function design, monotone Boolean functions are subject to strong structural constraints, which significantly limit the achievable nonlinearity. This makes them an interesting and challenging target for evolutionary search.

To address this problem, we considered three solution encodings: the standard truth table encoding, a balanced truth table encoding with dedicated variation operators, and a symbolic genetic programming representation. We also proposed fitness functions that combine nonlinearity optimization with explicit pressure toward monotonicity, and, in the balanced scenario, with an additional balancedness constraint. For imbalanced monotone functions, we investigated three variants of the monotonicity penalty in order to reduce the bias induced by Hamming weight.
The experimental study for dimensions $n=5,\dots,14$ shows that evolutionary algorithms can successfully discover monotone Boolean functions with high nonlinearity in both the balanced and imbalanced settings. In particular, the evolved functions consistently outperform majority functions in terms of nonlinearity, confirming that majority functions are not representative of the best nonlinear behavior achievable inside the monotone class. 
Overall, the results demonstrate that evolutionary search is a viable method for exploring the space of monotone Boolean functions with high nonlinearity. More specifically, TT tends to provide the strongest results in small and medium dimensions, whereas GP becomes more robust in the largest tested sizes, especially when TT fails to find feasible monotone solutions. This further indicates that the choice of representation and fitness design has a decisive impact on performance.

As future work, it would be natural to extend the search to larger dimensions, investigate additional representations and variation operators tailored to monotonicity, and analyze the structure of the best evolved functions in order to derive constructive principles for monotone Boolean functions with improved cryptographic properties.

\bibliographystyle{abbrv}
\bibliography{bibliography}

\end{document}